# On the Stability of Online Language Features: How Much Text do you Need to know a Person?

**Eben M. Haber**

IBM Research – Almaden, 650 Harry Road, San Jose, CA 95120
eben@acm.org

## Abstract

In recent years, numerous studies have inferred personality and other traits from people's online writing. While these studies are encouraging, more information is needed in order to use these techniques with confidence. How do linguistic features vary across different online media, and how much text is required to have a representative sample for a person? In this paper, we examine several large sets of online, user-generated text, drawn from Twitter, email, blogs, and online discussion forums. We examine and compare population-wide results for the linguistic measure LIWC, and the inferred traits of Big5 Personality and Basic Human Values. We also empirically measure the stability of these traits across different sized samples for each individual. Our results highlight the importance of tuning models to each online medium, and include guidelines for the minimum amount of text required for a representative result.

## Introduction

The idea of inferring personal traits from social media text has been very popular in recent years. This has particular relevance, since understanding people better from a readily available data source can permit us to build systems that better engage and meet the needs of users. While the results of trait-inferencing studies are encouraging – demonstrating correlations between inferred traits and the results of psychometric tests – several questions must be answered before we can use these techniques with confidence. How do these models vary across online media such as blogs, Twitter, and email? What is the minimum amount of text required to have a representative sample of a person's writing? Is temporal sample-size important?

In this paper, we address these questions by empirically studying several large sets of online, user-generated text, drawn from Twitter, email, blogs, and online discussions. We examine and compare population-wide results for the linguistic measure LIWC, and the inferred traits of Big5 Personality and Basic Human Values. We also empirically measure the stability of these traits across samples of different sizes and temporal ranges for each individual.

## Related Work

The field of psycholinguistics has a long history of understanding people through their language use. This, combined with the rise in popularity of social media, has led to research efforts into automatically understanding people through their online language and behavior, inferring traits such as personality (Gill 2011, Golbeck et. al. 2011 & 2011b, Yarkoni et. al. 2010), political orientation (Kosinski et. al. 2013), basic human values (Chen et. al. 2014), narcissism and psychopathy (Sumner et. al. 2012) and even post-partum depression (Choudhury et. al. 2014). While the reported accuracy of some of these predictions has been acceptable (e.g., Golbeck et. al. 2011, Gou et. al. 2014), none of these studies have reported how linguistic patterns differ across different online platforms, nor have they discussed how much text is required to have a representative sample for an individual.

## Experimental Setup

### Data Sources

To obtain a more complete picture of online linguistic patterns, we examined three distinct data sources: a corpus of Twitter data we collected, the Enron email corpus, and a collection of online community blog, forum, and wiki postings from a large enterprise.

Twitter is a popular social network, with more than 500 million postings per day. It is somewhat idiosyncratic, limiting posts to 140 characters, and by default making all posts publically visible. The public nature, combined with a free rate-limited API, make Twitter a valuable venue for research. As part of our investigations into individual online text patterns, for the past 2½ years we have been collecting temporally contiguous blocks of tweets for many individuals. We selected the individuals through a process similar to snowball sampling, collecting all public followers of more than 600 business and government accounts.

We collected tweets for more than 1.3 million of these followers. In most cases we collected only 200 tweets, but for more than 100k more active accounts we collected all available tweets. In our analysis of linguistic patterns, we only examine tweets in English, and we ignore re-tweets, since they don't represent that individual's language use.

As a counterpoint to the social, idiosyncratic language of Twitter, the Enron email corpus provides a large example of business e-mail (Cohen) from early in the 21$^{st}$ century. Made public during litigation over the failure of the Enron corporation, it includes the email folders of 150 Enron employees, with approximately 500k messages. When the messages are sorted by sender, it is possible to find 527 distinct people (as opposed to automated email accounts) who wrote at least 100 messages with a total of at least 1000 words of text overall. In our analysis, we exclude all quoted text in replies and forwarded messages, since that represents another person's language.

Our third source of online text comes from the blogs, forums, and wiki edits in more than 1000 public online communities at a large enterprise. Crawled in 2014, this contains all public contributions of more than 46,000 employees, though only those with at least 5000 words of text were analyzed (1800 employees).

**Linguistic Measures**

To analyze the text, we chose three measures. The first is the LIWC 2001 dictionary, a widely used tool that assigns common English words to one of 68-different categories (e.g., pronouns, adverbs, words about work, words about religion, etc.) (Pennebaker et. al. 2007). LIWC is widely used, and provides an objective summary of the types words in use.

We also include two higher-level models that infer individual traits from language use. The first, described by Yarkoni, predicts the Big5 personality traits and 30 sub-facets. The second (Chen et. al. 2014), predicts Basic Human Values (BHV), the motivational factors that influence an individual's decisions about the future (Schwartz 2006). Both of these models are based on LIWC 2001.

## Results

**Population-Wide Results**

The first question we examine is how these models compare across different online media. Yarkoni built his model using text from blogs, and Chen used text from Reddit, so it is important to see how they vary with text from varied sources.

Using the data sources listed above, we collected at least 5000 words of text from individuals writing in email (n=485), blogs (n=518), forums (n=443), wikis (n=799), and Twitter (n=263,165), and computed the mean values for all traits from LIWC, Yarkoni's Big5 model, and Chen's BHV model. The differences in LIWC alone are quite dramatic. Considering only blogs, forums, and wikis, which were all drawn from a single enterprise's internal online communities, 10/68 LIWC traits showed significant ($p < 0.001$) and large ($> 0.8sd$) differences in the means. Figure 1 shows the relative mean frequencies of these LIWC word categories across media. Some differences are not surprising, e.g., the low usage of pronouns or references to others in wikis compared to forums and blogs. Other differences, such as discrepancy, present, and exclusion words are less intuitive, but still large, e.g., discrepancy words are nearly twice as common in forums than blogs, two media of enterprise communication that one would expect to be more similar.

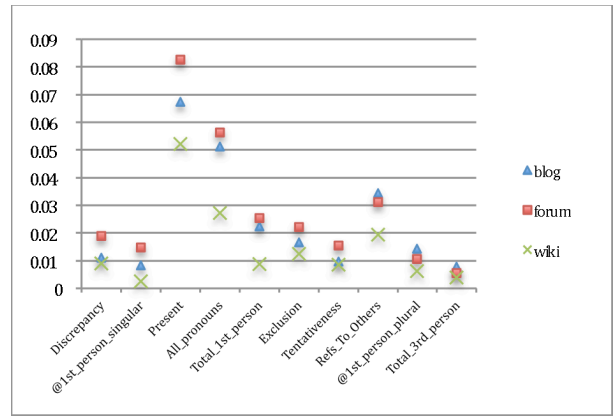

**Figure 1. Frequency of certain LIWC categories in blogs, wikis, and forums.**

The differences get more extreme when comparing the Twitter, email, and enterprise data. Now, 54/68 categories show large differences ($> 0.8sd$), the most extreme being profanity, which is 70x more common on Twitter than on the other, more business oriented media, along with words about sleep (29x), religion (19x), sexuality (14x). Of course several categories occur less often on Twitter: words about jobs (0.4x), numbers (0.56x), occupations (0.57x), school (0.58x), and 1$^{st}$ person plural (e.g. we, us, 0.72x).

Most of these differences make sense, but even when their cause is unclear, they are important because Yarkoni and Chen's models use these word categories as significant predictors in their models. Given the variation in word category frequency between these online media, we expect similar variation in the models based on that word use. This is visible in Figure 2, which compares the ranges of Big5 and BHV traits across different media. For each trait, its 95% confidence interval is shown for Twitter, email, blogs, forums, and wikis, from left to right. The y-axis is normalized to put the mean for Twitter at 1. Given the variation in these ranges, it is clear that the values of these traits inferred from one online medium are often not comparable with another. E.g., a high value of the Big 5 Open-ness trait inferred from twitter would be at the low end of the range for wikis, with the opposite true for Extraversion. Gou 2014 and Chen 2014 both use models trained in one domain (blogs or Reddit) applied to another domain (Twitter); these results suggest their approach potentially introduced error, and that they should have performed renormalization based on the word category frequencies in each medium. In general, it is clear that language samples and inferred traits from one medium should not be compared with those from another medium without renormalization.

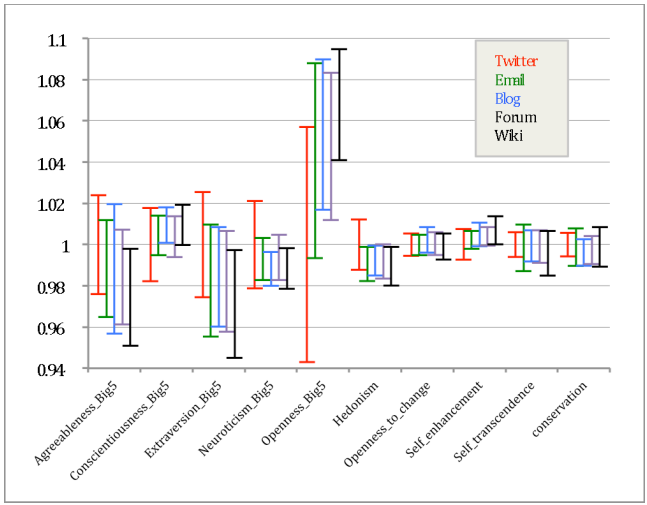

**Figure 2. 95% confidence intervals for Big 5 and basic human values traits inferred from various online media.**

## Variability – how much text is enough?

The second question we aim to answer relates to sample size: if linguistic patterns are indicative of individual traits, how much text is required to get a representative sample of those patterns? Is a minimum temporal sample size important? Sample-to-sample variations put a ceiling on predictive accuracy - even the most accurate model can be no more precise than the variations in component features.

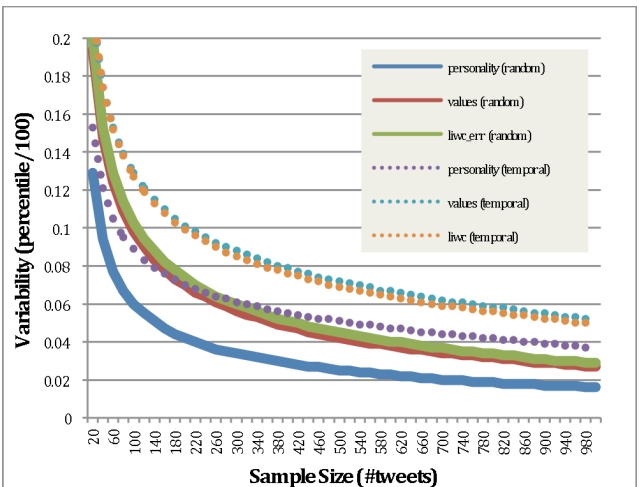

**Figure 3. Variability for Subsamples of 20-1000 Tweets**

We decided to measure this empirically, for people with large amounts of text, comparing linguistic measures on the whole of an individual's text to the same measures for various sized subsamples. We ran this experiment separately for each of the data sources above, and we computed and compared the values for all LIWC, personality, and values traits. Sample size was computed either by message counts, or by word counts, for which we tried different threshold values. For example, we tried a threshold value of 100 email messages on the Enron corpus. Thus, we took each person with at least 100 messages, computed the linguistic measures for those 100 messages, then compared that with the measures computed for 20 subsamples of 5 messages, 10 samples of 10 messages, and so on up to two samples of 50 messages. We tried both randomly selected subsamples, and temporally contiguous subsamples.

Since trait values fall in a variety of ranges, we normalized all the traits into population percentiles, and report variability in terms of percentile. E.g., if the measure of Extraversion for 100 messages placed a person in the $35^{th}$ percentile for the population, and the measure for a subsample of 10 messages placed that person in the $50^{th}$ percentile, then the variability for that subsample would be 15 percentile. This has the effect of magnifying differences in the middle of the range, and minimizing differences on the tails, but we deemed this acceptable in order to compare variability across all traits in an intuitive manner. We measured both average and std. dev. of this difference, so we can report both the average variability across all people for a subsample size, and also the 95% confidence that variability will be below a certain level.

Figure 3 shows average variability for Twitter data organized by number of tweets, where the total sample size for each person was 2000, and subsample sizes ranged from 20 to 1000 tweets. Note that: 1) Variability is notably higher for temporally contiguous subsamples as compared to randomly selected subsamples. This suggests that messages close together in time are more similar to each other and vary more from the whole than widely separated messages. Thus, temporal sample size is important in achieving a stable measure, further study will be required to determine the time period required. 2) The Big 5 model shows lower variability than the values model, though both are based on LIWC. This suggests that some models are more resistant to noise/error that others. 3) Even for temporally-contiguous samples, only 200 Tweets is required to obtain a mean variability of less than +/- 10 percentile, a relatively small amount of text.

A separate experiment examining a model that incorporated both bag-of-words-based and LIWC found much higher variability, not surprisingly. This suggests that variability is an important metric to measure when evaluating new models.

How much text is enough? Of course that depends on the required precision, and whether an average or absolute measure of variability is required. Figure 4 shows the result of Big5 variability using word-count-based samples for people with at least 15,000 words. It shows, for each word count sample size, both the average variability, and 95% confidence maximum variability. Twitter shows the least variability for any given sample size, probably because of its short message size, and email the most, because of its longer message size. In general, average variability is less than +/-10 percentile for a sample of 4000-5000 words, though a much larger sample is required to be 95% confident of a low variability.

## Discussion and Conclusion

In this paper we analyzed large amounts of real-world online text from Twitter, email, blogs, forums, and wiki edits. We demonstrated that standard linguistic measures of these media differ dramatically, no doubt due to stylistic and topical differences between media, suggesting that renormalization and tuning is required to move a model from one medium to another. We also obtained estimates of the amount of text required to get a representative sample, which also differs across media.

This work is not intended to be the last word on these issues, indeed it has many limitations. The Enron corpus is more than a decade old, yet it is still unique in its size and scope. Our collection of Twitter data has been oriented towards quantity over quality, we have not found a good way to validate or categorize the million+ accounts in our data set, but believe the sheer quantity gives us a reasonable sample. Finally, we have not been able to collect data to compare the same people across different media – even with our enterprise data too few people post sufficient text across multiple media. Even so, we were able to show that linguistic variation across online media, even intuitively similar media like enterprise blogs and forums, is sufficiently large to be a confounding factor in moving models between domains, that temporal sample-size is important, and variability is a useful feature in determining an appropriate sample size for analysis. We hope this paper inspires future investigations of linguistic variation on and across other data sets, so that social media linguistic models can become more accurate and useful.

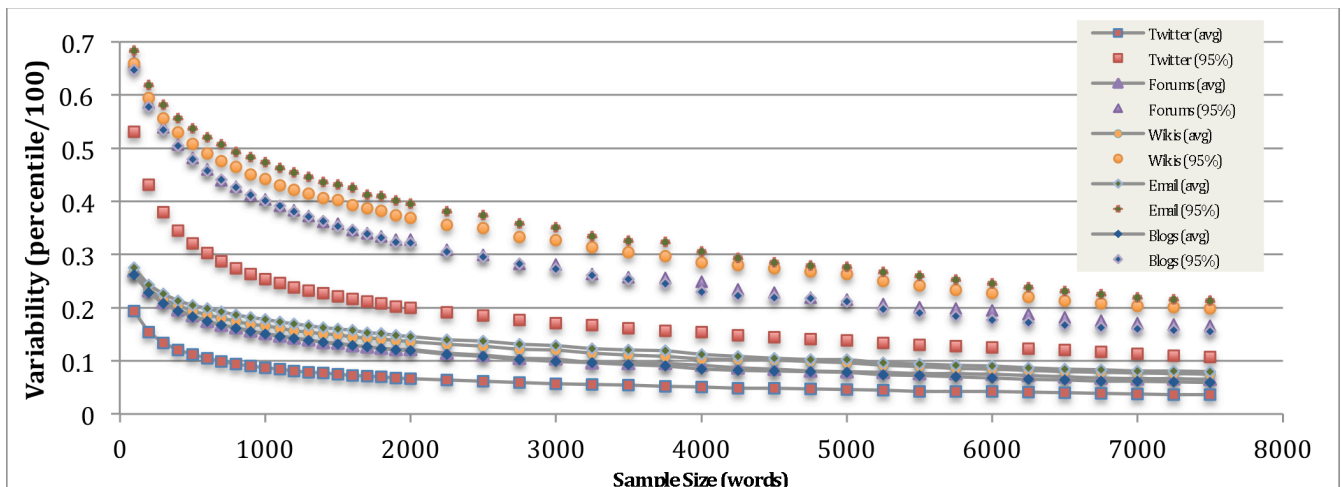

**Figure 4. Variability for Personality, By Word Count, for Different online media.**